\documentclass[10pt,twocolumn,letterpaper]{article}

\usepackage{iccv}
\usepackage{times}
\usepackage{epsfig}
\usepackage{graphicx}
\usepackage{amsmath}
\usepackage{amssymb}
\DeclareUnicodeCharacter{2212}{-}%fixes unicode error

\usepackage{booktabs}
\usepackage{algorithm}
\usepackage{algpseudocode}
\usepackage{censor}
\usepackage[numbers,sort&compress]{natbib}

\newcount\Comments  % 1 suppresses notes to selves in text
\usepackage{color}
\definecolor{darkgreen}{rgb}{0,0.5,0}
\newcommand{\kibitz}[2]{\ifnum\Comments=0\textcolor{#1}{#2}\fi}

\usepackage{hyperref}
\hypersetup{colorlinks,allcolors=black}

\iccvfinalcopy % *** Uncomment this line for the final submission

\def\iccvPaperID{10884} % *** Enter the ICCV Paper ID here

\ificcvfinal\fi

\begin{document}

%%%%%%%%% TITLE
\title{Localization-based Tracking}
\author{Derek Gloudemans and  Daniel B. Work\\
Institute for Software Integrated Systems\\
Vanderbilt University\\
1025 16th Avenue S., Nashville, TN 37212\\
{\tt\small derek.gloudemans@vanderbilt.edu}
}

\maketitle
\ificcvfinal\thispagestyle{empty}\fi

%%%%%%%%% ABSTRACT
\begin{abstract}
   End-to-end production of object tracklets from high resolution video in real-time and with high accuracy remains a challenging problem due to the cost of object detection on each frame. In this work we present \textit{Localization-based Tracking} (LBT), an extension to any tracker that follows the tracking by detection or joint detection and tracking paradigms. Localization-based Tracking focuses only on regions likely to contain objects to boost detection speed and avoid matching errors. We evaluate LBT as an extension to two example trackers (KIOU and SORT) on  the UA-DETRAC and MOT20 datasets. LBT-extended trackers outperform all other reported algorithms in terms of PR-MOTA, PR-MOTP, and mostly tracked objects on the UA-DETRAC benchmark, establishing a new state-of-the art. Relative to tracking by detection with KIOU, LBT-extended KIOU achieves a 25\% higher frame-rate and is 1.1\% more accurate in terms of PR-MOTA on the UA-DETRAC dataset. LBT-extended SORT  achieves a 62\% speedup and a 3.2\% increase in PR-MOTA on the UA-DETRAC dataset. On MOT20, LBT-extended KIOU has a 50\% higher frame-rate than tracking by detection and is 0.4\% more accurate in terms of MOTA. As of submission time, our LBT-extended KIOU tracker places 10th overall on the MOT20 benchmark.
\end{abstract}

\section{Introduction}

In this work we address the task of \textit{multiple object tracking} (MOT) from raw video sequences. The goal of this task is to detect the bounding box of each object at each frame in the video, and to associate these bounding boxes across frames to provide matched identities for each unique object across time. These detections, matched across all frames to identify unique objects, are referred to as \textit{tracklets}. The real-time performance of object detection and tracking is paramount in a number of domains, such as autonomous vehicle sensing \cite{geiger2012we}, robotics \cite{chen2011kalman,wang2007simultaneous}, traffic management \cite{alexiadis2004next,krajewski2018highd}, automated surveillance \cite{del2012efficient}, smart city management \cite{hu2017iot,hancke2013role} and augmented reality systems~\cite{wagner2009real}.

\begin{figure}
    \centering
    \includegraphics[width = \columnwidth]{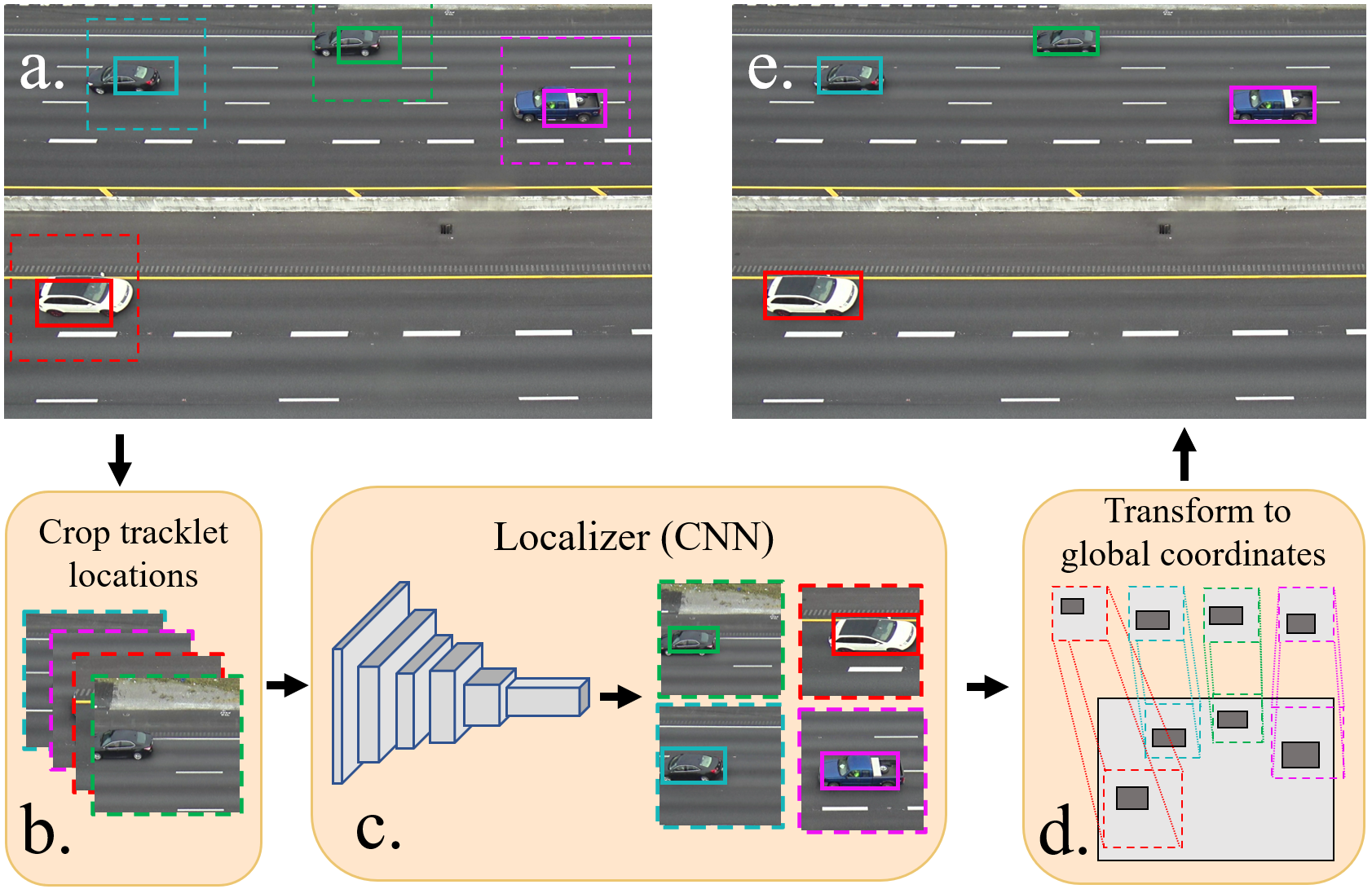}
    \caption{Overview of Localization-based Tracking (proposed). \textbf{(a)} Tracklet \textit{a priori} locations (solid boxes) are used to \textbf{(b)}. crop (dashed boxes) around likely object locations. \textbf{(c)}. These crops are processed by the localizer to localize each object. \textbf{(d)}. The resulting bounding boxes (solid boxes) are then transformed back into the coordinate system of the frame, \textbf{(e)}. producing final object detections for each tracklet.}% All un-cropped portions of the image are ignored.}
    \label{fig:overview}
\end{figure}

The vast majority of algorithms for multiple object tracking decompose the problem into two distinct tasks: First, the \textit{object detection} task locates relevant objects within a frame. Second, the \textit{object association} (commonly referred to simply as object tracking) task associates or matches objects in the current frame with the same objects in the previous frame such that each object is uniquely identified across the entire video sequence. Importantly, though, tracking by detection methods are hampered by the fact that the best-performing algorithms in terms of detection accuracy still run below 30 frames per second on a GPU for frames of modest size (e.g., 960$\times$540 \cite{lyu2018ua}, 1392$\times$512 \cite{geiger2012we}, and 1920$\times$1080 \cite{MOT19_CVPR,MOTChallenge20}. A variety of recent methods \cite{feichtenhofer2017detect,pang2020tubetk,sun2020simultaneous, zhou2020tracking,bergmann2019tracking,li2020smot,wang2019towards,peng2020chained} have sought to leverage the tracking context to provide additional information for the task of object detection, performing detection and tracking \textit{jointly} rather than in parallel. These methods make use of additional information to boost object tracking accuracy rather than speed. Unfortunately, the increased accuracy is not realizable if real-time requirements must be met, which is true for many domains where object tracking is used.

In this work we introduce  \textit{Localization-based Tracking} (LBT) to increase the speed of object detection and tracking. We make use of the tracking context to inform and speed object detection. Intuitively, object locations in previous frames (possibly complemented with a motion model) provide a strong prior for object locations in the current frame. Our method seeks to leverage this intuition to reduce the search space to detect the object in the next frame. A graphical overview of LBT is given in Figure \ref{fig:overview}. To make use of object priors from previous frames (a.),  we  (b.) crop portions of the frame  we expect to contain existing tracked objects. (c.) Within each crop, we solve the task of \textit{localization} (identifying the location of a single object of interest) with a \textit{localizer} (an object detector with additional logic for selecting the best bounding box from amongst candidate bounding boxes). LBT is a general approach and can be used to extend base object trackers to improve speed and accuracy with minimal additional work.

\begin{figure}
    \includegraphics[width = 0.9\columnwidth]{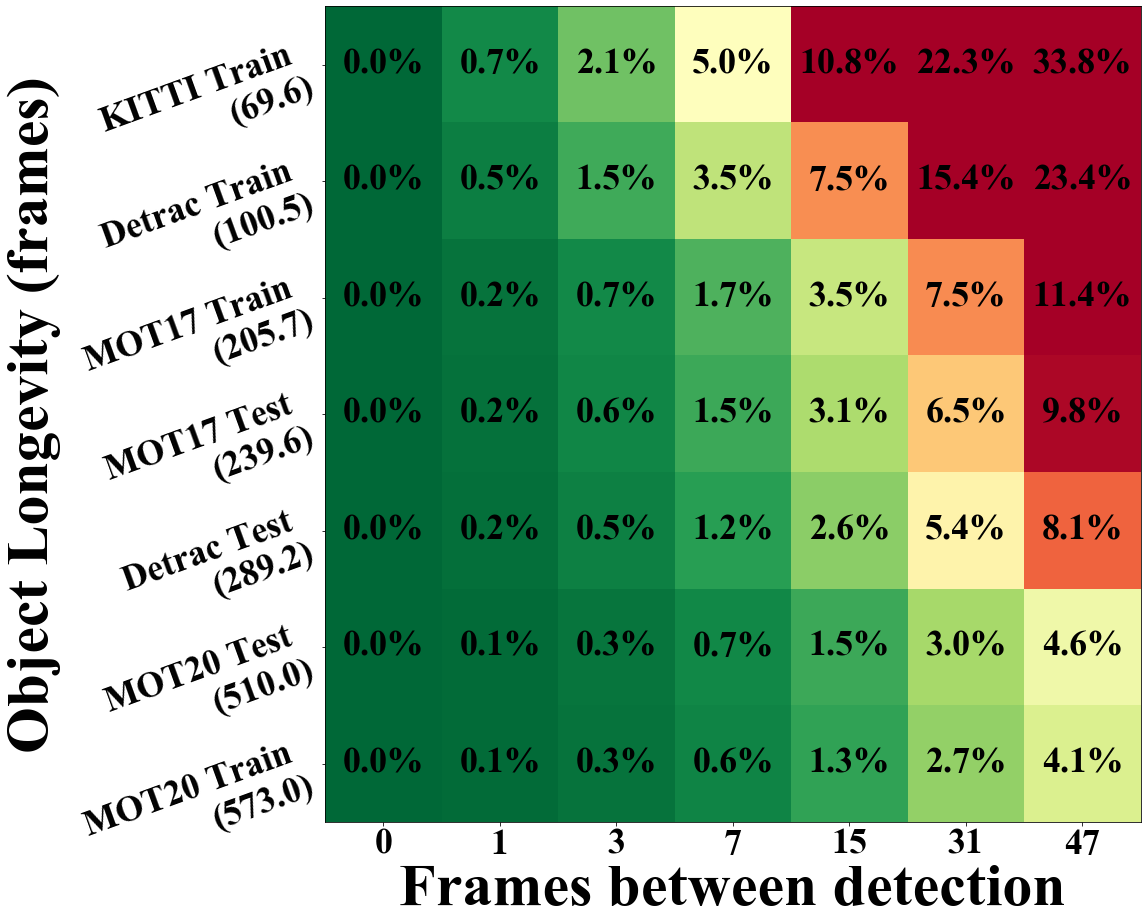}
    \caption{Expected increase in false negative rate (false negatives as a fraction of total ground-truth objects) due to new undetected objects (false negatives) appearing between detection steps on various datasets. When detections occur every $d$ frames, on average a new object is missed in $\frac{d}{2}$ frames after it initially enters the frame.}
    \label{fig:heatmap}
\end{figure}

In LBT, because detection is not run on the full frame for every frame, there is a potential to increase the number of false negatives due to missed detections when objects first appear. However,  Figure \ref{fig:heatmap} shows that the expected increase in the false negative rate is small for many real-world datasets. Moreover, we show in Section \ref{sec:results} that LBT improves overall tracking performance because of a dramatic reduction in false positives that appear when performing detection on every frame. 

The main contribution of this work is as follows. We introduce the Localization-based Tracking framework, a general extension to existing trackers which explicitly leverages the tracking context to achieve state-of-the-art performance. We use LBT to extend two example trackers (KIOU and SORT) and we evaluate the extended trackers on two common multiple object tracking benchmarks, UA-DETRAC and MOT20. On UA-DETRAC, LBT-extended trackers outperform all existing methods in terms of PR-MOTA, PR-MOTP, and mostly tracked objects. Moreover, LBT increases speed and in most cases accuracy relative to the base tracker on both benchmarks. Specifically, KIOU-based LBT on UA-DETRAC achieves a 25\% speedup and a +3.2\% PR-MOTA; SORT-based LBT on UA-DETRAC achieves a 62\% speedup and +1.1\% PR-MOTA; SORT-based LBT on MOT20 achieves a 50\% speedup and +0.4\% MOTA; and SORT-based LBT on MOT20 achieves a 78.8\% speedup with -4.8\% MOTA.

The remainder of this article is organized as follows. In Section \ref{sec:related},  existing approaches to multiple object tracking are reviewed. Section \ref{sec:methodology} introduces the Localization-based Tracking extension and describes its implementation. Section \ref{sec:experiments} details the experiments used evaluate LBT, and Section \ref{sec:results} presents the results of these experiments. Section \ref{sec:future} concludes the article by identifying promising directions for future research on fast multi object tracking.

\section{Background}
\label{sec:related}
Multiple object tracking algorithms such as~\cite{bewley2016simple,wojke2017simple,bochinski2017high,bochinski2018extending,kiout,milan2013continuous,zhan2020simple,bergmann2019tracking,xu2019deepmot,karthik2020simple,xiang2015learning,chu2017online,yan2012track,feichtenhofer2017detect,pang2020tubetk,sun2020simultaneous,zhou2020tracking,wang2019towards,peng2020chained,li2020smot} can be roughly divided into three main categories: tracking by detection methods, integrated object detection and object association, and adaptations of single object trackers. The first two categories are extensible using LBT. We briefly review prominent methods in each category. 

\noindent \textbf{Tracking by Detection}. Most modern object trackers follow the tracking by detection framework~\cite{ fan2016survey,luo2014multiple} This paradigm divides the task of producing tracked objects from raw video into two steps. First, relevant objects are detected in each frame. Then, detected objects are associated across frames \cite{bochinski2017high}. This subdivision of tasks into detection and tracking is encouraged by popular benchmarking datasets for object detection and tracking such as MOTChallenge~\cite{milan2016mot16}, KITTI~\cite{geiger2012we}, and UA-DETRAC~\cite{wen2015ua}.

\noindent \textbf{Object Detection}. Most top-performing object detectors rely on \textit{convolutional neural networks} (CNNs) for feature extraction from an image. Common approaches include one stage detection, such as in YOLO \cite{redmon2016you}, RetinaNet \cite{Lin_2017_ICCV} and SSD \cite{liu2016ssd}, where features output by a convolutional neural network are directly used to regress object bounding box coordinates. Two stage detectors such as Faster RCNN \cite{ren2015faster} and Evolving Boxes \cite{wang2017evolving} attempt to boost accuracy by adding an intermediate step in which promising candidate regions with a high probability of being objects are selected, and then only from these regions are regressed bounding boxes in the second stage. Segmentation models such as Mask-RCNN \cite{he2017mask}, and deep watershed transform-based segmentation \cite{bai2017deep} have also been adapted to perform bounding box-based detection. Recently, these approaches have been bolstered by adding additional awareness of foreground and background \cite{fu2019foreground}, by use of attention networks~\cite{perreault2020spotnet,wu2019hierarchical}, or by regressing the location of keypoints such as bounding box corners \cite{law2018cornernet} or object centers \cite{duan2019centernet} with custom pooling layers that better convey keypoint information through convolutional layers. Importantly, though, the best-performing algorithms in terms of detection accuracy still run below 30 frames per second on a GPU for frames of modest size (e.g., 960$\times$540 \cite{lyu2018ua}, 1392$\times$512 \cite{geiger2012we}, and 1920$\times$1080 \cite{MOT19_CVPR,MOTChallenge20}).  

\noindent \textbf{Object Association}. Object association methods compare objects from sequential frames in terms of position, appearance, and/or physical dynamics to match objects from one frame to the next.  In \textit{Simple Online Realtime Tracking} (SORT) \cite{bewley2016simple}, Kalman filtering is used to predict object locations and these predicted positions are matched to current frame detections based on straight-line distance. DeepSORT \cite{wojke2017simple} refines SORT by additionally using an appearance embedding for each object to aid matching. The IOU tracker \cite{bochinski2017high} utilizes bounding box overlap rather than straight-line distance as the distance metric, and K-IOU \cite{kiout} combines this method with Kalman filtering for more accurate matching. \textit{Continuous Energy Minimization} (CEM) combines bounding box overlap ratio, color-based appearance dissimilarity, object physical dynamics, and logical constraints to match objects \cite{milan2013continuous}. Other successful recent approaches for cross-frame association leverage CNN-based reID features and \cite{zhan2020simple,zhang2020fairmot,liang2020rethinking} or otherwise explicitly regress future object positions from a frame using CNNs \cite{bergmann2019tracking}. Neural network based trackers are shown to benefit from directly differentiable tracking-specific loss functions in \cite{xu2019deepmot}. Weakly supervised methods such as SimpleReID \cite{karthik2020simple} have also been successful for tracking tasks. 

\noindent \textbf{Single Object Tracking Methods for MOT}. Recently, a few works have approached multiple object tracking as a set of parallel \textit{single object tracking} (SOT) tasks. In the SOT problem, a single object is manually identified in a frame and is subsequently tracked. Extremely fast algorithms using kernel-based methods \cite{hare2015struck} and regression networks \cite{held2016learning} have been developed. The parallel SOT task has been posed as a Markov Decision Process \cite{xiang2015learning}, or handled using a CNN with single-target-specific branches that utilize shared features \cite{chu2017online}. The work proposed in \cite{yan2012track} combines both object detector detections and rough single object tracking object positions to refine object position estimates. Similarly, VIOU \cite{bochinski2018extending} extends IOU tracking by employing a single object tracker to recover lost objects, reducing fragmentations. Thus far, SOT-based methods for multiple object tracking have not been scalable in terms of speed, due either to  \textit{i.}) the need to update appearance models for each object online or \textit{ii.}) the need to initialize new objects, which has been addressed by additionally performing detection on each frame \cite{he2017sot}.

\noindent \textbf{Joint Detection and Tracking.}
Very recently, a number of works have achieved high accuracy on multiple object tracking by integrating information from the tracking context for object detection. Some approaches pass pairs \cite{feichtenhofer2017detect} (or larger sequences \cite{pang2020tubetk,sun2020simultaneous}) of consecutive frames to CNNs to regress ``tubelike'' detections across multiple frames (a matching step is still required). \cite{zhou2020tracking} passes the previous frame as well as a "heatmap" of previous object positions  to the object detector which predicts object locations and offsets used to aid greedy matching of objects.  Other works perform tracking by object re-detection, passing previous inputs to an object detector as additional anchor boxes \cite{bergmann2019tracking,li2020smot}.  \cite{wang2019towards} uses the same CNN to output bounding box coordinates and object embeddings for re-identification, and uses a fast matching approach to associate objects based on these embeddings. \cite{peng2020chained} regresses bounding box pairs for consecutive frames, performing IOU-based matching on the two sets of bounding boxes corresponding to each single frame. Recently, graph neural networks have also been used to leverage spatial-temporal relationships for both object detection and object association~\cite{wang2020joint}.

Most of the trackers in this category pass information on tracklet locations to the detection step to refine object detection and boost accuracy, rather than to increase the speed of object detection. While our approach is most similar to these integrated object detection and tracking approaches \cite{bergmann2019tracking,li2020smot}, the primary focus of our tracker extension is on faster rather than more accurate performance. The approach proposed in this work is also somewhat similar to SOT methods for multiple object tracking \cite{bochinski2018extending}, but explicitly outputs multiple objects to prevent switches in crowded settings, and allows any detector and tracking framework to be used for localizing a single object. 

\section{Methodology}
\label{sec:methodology}
Localization-based Tracking is an enhancement to existing tracking methods. It can be utilized with any multiple object tracking scheme that makes use of an object detection step to detect multiple objects. In the following section, we provide a high-level overview of LBT, and extend two tracking by detection methods to LBT as  illustrative examples.%  then explain how to extend a tracker with LBT.

\subsection{Overview}
\begin{figure}
    \centering
    \includegraphics[width = 0.9\columnwidth]{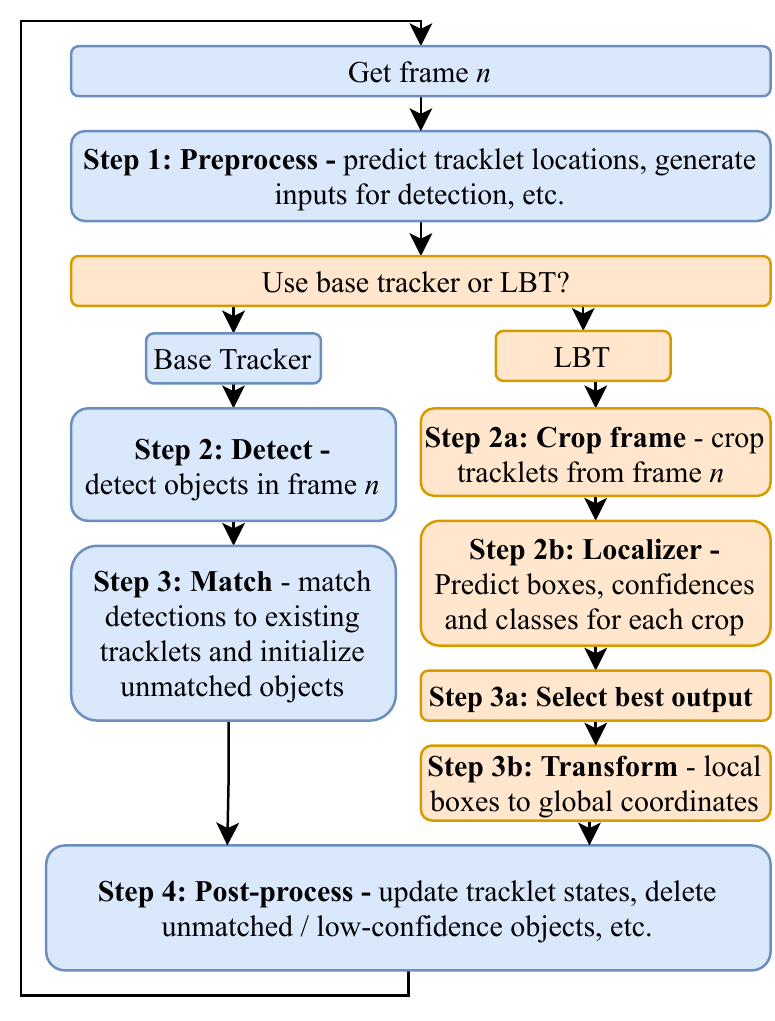}
    \caption{Tracking by localization (orange) extending base tracker (blue).}
    \label{fig:LBT-overview}
\end{figure}

Localization-based Tracking extends a base tracker to jointly produce object detections and tracks for a video sequence in an online rather than batched fashion, utilizing information from the tracking context to aid and speed detection. Rather than performing (slow) object detection every frame, LBT skips $d$ frames before performing object detection. On detection frames, the base tracker is used to associate detections with existing object tracks, and to initialize and remove tracked objects. On all other frames, existing object tracklets are updated using localization.  Specifically, the estimated location of each existing object is used to crop a portion of the frame around that object within the the overall frame. These small crops are then processed by a \textit {localizer}, which is a specialized detector that localizes a single object within a crop. All portions of the image that are not already associated to a tracklet are ignored. Because each crop is already associated to an object when it is generated, there is no need to perform a (possibly error prone) data association step. Figure \ref{fig:LBT-overview} provides a graphical overview of a tracker extended with LBT. 

\subsection{Extending a tracker with LBT}
The proposed method builds on the standard process for multi-object tracking:

    \noindent\textbf{Step 1: Preprocess.} Use tracklets from frame $0,\cdots$, $n-1$ to generate estimated \textit{a priori} object locations in frame $n$.% such as estimated \textit{a priori} object locations.
    
    \noindent\textbf{Step 2: Detect.} Produce detections for frame $n$. Optionally, also produce information for matching detections in frame $n$ to tracklets.
    
    \noindent\textbf{Step 3: Match.} Associate detections from frame $n$ to existing tracklets. Initialize new tracklets for unmatched detections.
    
    \noindent\textbf{Step 4: Postprocess.} Generate \textit{a posteriori} object locations at $n$; remove unmatched tracklets.

To extend such a tracker with Localization-based Tracking, steps 2 and 3 above are replaced with the following process:% Figure \ref{fig:LBT-overview} shows a graphical summary of this new process: 

\begin{figure}
    \centering
    \includegraphics[width = 0.9\columnwidth]{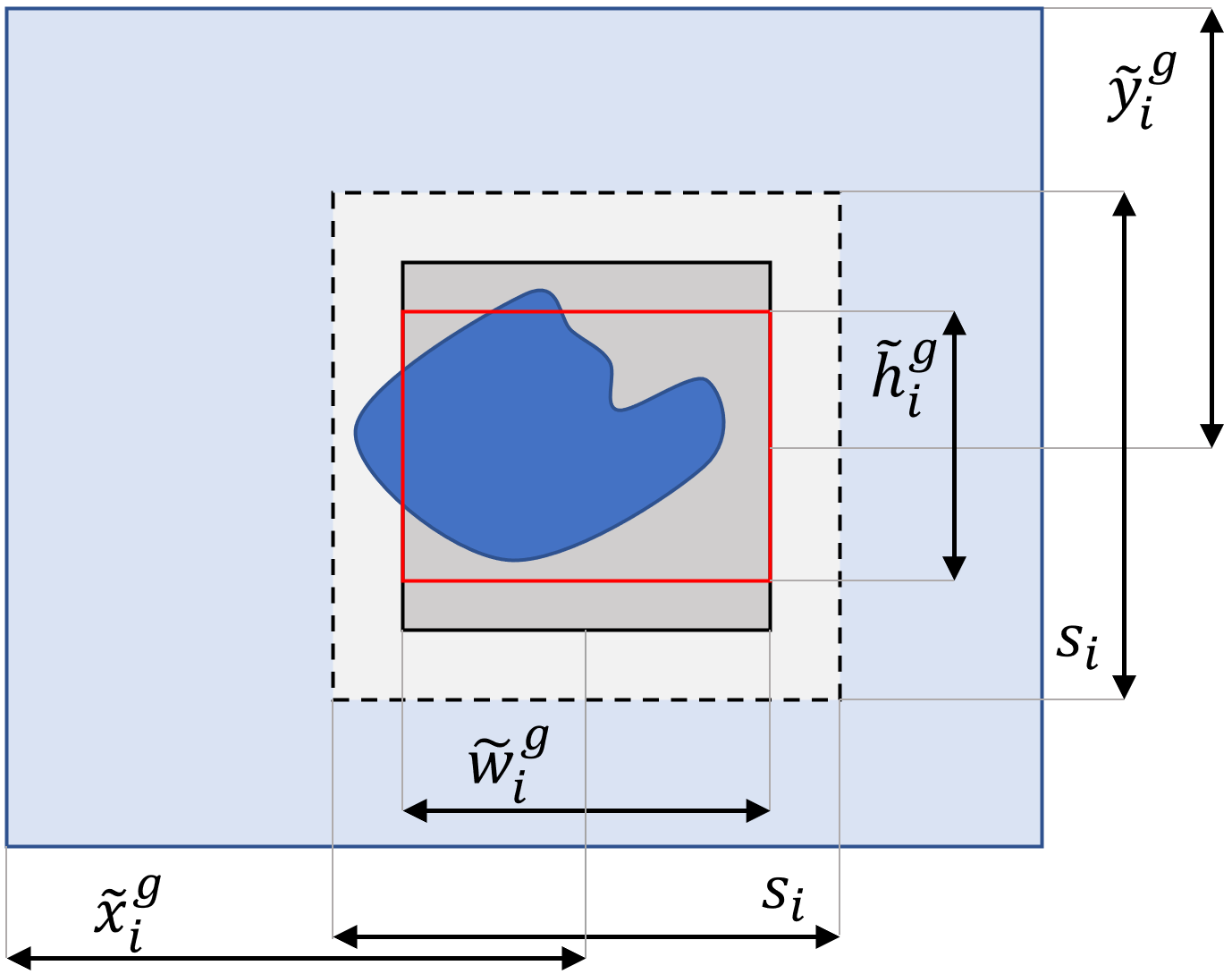}
    \caption{\textit{a priori} object $i$ location in global (frame) coordinates, $\text{b}\tilde{\text{o}}\text{x}^g_{i} = [\tilde{x}^g_{i},\tilde{y}^g_{i},\tilde{w}^g_{i},\tilde{h}^g_{i}]$ (red), is made square (black solid) and expanded by a factor of $\beta$ to produce $\text{crop}_i := [\tilde{x}^g_{i},\tilde{y}^g_{i},s_{i}]$ (black dash) before being passed to the localizer. Expansion helps to ensure that actual tracked object (dark blue) will be contained within the crop area.}
    \label{fig:crop}
\end{figure}

\noindent\textbf{Step 2a: Crop frame.} Generate a square cropping box centered on each \textit{a priori} object location in frame $n$. To ensure that the full object is contained within the crop, expand the crop to be larger than the a priori object estimate.% somewhat larger than the object. 

More precisely, let $\mathcal{O}:=\{1,\cdots,i,\cdots,o_{max}\}$ be the set of all tracked objects, indexed by $i$. The \textit{a priori} (denoted by tilde) bounding box for object $i$ in global (frame) coordinates is defined by the center x-coordinate $\tilde{x}^g_{i}$, the center y-coordinate $\tilde{y}^g_{i}$, the width $\tilde{w}^g_{i}$ and height $\tilde{h}^g_{i}$. Let $\text{b}\widetilde{\text{o}}\text{x}^g_{i}: = [\tilde{x}^g_{i},\tilde{y}^g_{i},\tilde{w}^g_{i},\tilde{h}^g_{i}]$.  Similarly, we define the corresponding square crop for object $i$ as $\text{crop}_i := [\tilde{x}^g_{i},\tilde{y}^g_{i},s_{i}]$, where $s_{i}$ is the scale. The scale is computed as:
\begin{equation}
    s_{i}=\max\{\tilde{w}^g_{i},\tilde{h}^g_{i}\} \times \beta,
\end{equation}
where $\beta$ is a box expansion ratio (a parameter) used to ensure the full object is within the crop. Figure \ref{fig:crop} shows a graphical representation of crop generation for a single object.

 %   \begin{equation}
 %   \begin{split}
 %       x^c_{i} & = \tilde{x}^g_{i} \\
 %       y^c_{i} & = \tilde{y}^g_{i} \\
 %       s^c_{i} & = \max(\tilde{w}^g_{i},\tilde{h}^g_{i}) \times \beta, \\
 %   \end{split}
 %   \end{equation}
  %  where $\beta$ is the box expansion ratio, a tuned parameter. Figure \ref{fig:crop} shows a graphical representation of crop generation for a single object.

By construction, $\text{crop}_i$ is of size ($s_{i} \times s_{i}$) pixels. Before passing to the localizer, each crop re-scaled to a size ($C\times C$) pixels, where $C$ is a constant across all crops.  Other relevant feature maps such as previous frames \cite{feichtenhofer2017detect,pang2020tubetk,sun2020simultaneous,peng2020chained} or object center heat-maps \cite{zhou2020tracking} can be similarly cropped at this stage.
    
\noindent\textbf{Step 2b: Localize.}  All image crops corresponding to \textit{a priori} object locations are processed by the localizer, which produces bounding boxes that estimate the location of the  object within each crop. The localizer is identical in structure to the detector utilized by the base tracker, but is trained separately from the detector on images of size ($C \times C$), due to the difference in scale. Object detectors (and thus the localizer) inherently produce a number of bounding box outputs. Given $\text{crop}_i$, the localizer returns $l_{max}$ bounding boxes indexed by $j$ in the local crop coordinates. Each localizer output is an estimated location of object $i$ within the crop, defined by the object center, box width, and box height. The $j$-th bounding box output of the localizer corresponding to $\text{crop}_i$ is written as $\text{box}^l_{i,j} := [x^l_{i,j},y^l_{i,j},w^l_{i,j},h^l_{i,j},\text{conf}_{i,j}]$, where $\text{conf}_{i,j}\in[0,1]$ is the confidence of the $j$-th localizer output associated with $\text{crop}_i$.

\noindent\textbf{Step 3a: Select localizer output.} The best bounding box for each crop is selected from among the localizer outputs for the corresponding crop.  We use the same distance metric as the base tracker for matching in Step 3 of the base tracker, with the caveat that object confidence outputs from the localizer are also taken into account.  A detailed example is provided in Section \ref{sec:setup}. The best localizer output corresponding to $\text{crop}_i$ is written as $\text{box}^l_{i} := [x^l_{i},y^l_{i},w^l_{i},h^l_{i}]$, in coordinates local to the crop. Since the set of localizer outputs for a crop are compared to the single \textit{a priori} object $i$'s location, output selection across all objects is $\mathbf{O}(o_{max} \times l_{max})$ in complexity, where $o_{max}$ is the total number of tracked objects and $l_{max}$ is the total number of localizer outputs per crop. This operation is significantly less complex than a $\mathbf{O}(o_{max}^3)$ global min-cost matching problem in Step 3 of the base tracker \cite{ford2015flows}. Moreover it avoids the corresponding object association errors that can occur in Step 3. 

\noindent\textbf{Step 3b: Local to global transformation.} The best localizer output $\text{box}^l_{i}$ for each crop $i$ is converted back into global coordinates, where it can be used to update the $i$-th tracklet. In global coordinates, $\text{box}^l_{i}$ is denoted as  $\text{b}\hat{\text{o}}\text{x}^{g}_{i} := [\hat{x}^g_{i},\hat{y}^g_{i},\hat{w}^g_{i},\hat{h}^g_{i}]$, where:
\begin{equation}
\begin{split}
    \hat{x}^g_{i} & = x^l_{i} \times \frac{s_{i}}{C} + \tilde{x}^g_{i}, \\
    \hat{y}^g_{i} & = y^l_{i} \times \frac{s_{i}}{C} + \tilde{y}^g_{i}, \\
    \hat{w}^g_{i} & = w^l_{i} \times \frac{s_{i}}{C}, \\
    \hat{h}^g_{i} & = h^l_{i} \times \frac{s_{i}}{C}.
\end{split}
\end{equation}
Step 4 is then performed as for the base tracker. Each $\text{b}\hat{\text{o}}\text{x}^g_{i}$ is used to update the tracklet for object $i$.

\subsection{Adding new objects} Periodically, frames are processed by the detector rather than the localizer, so that new objects can be detected. On these frames, the base tracker is used to match detections to existing tracklets. Any detections without a suitable match to an existing tracklet are initialized as new tracklets. The number of frames between detection $d$ is variable, and can be tuned according to the context of the video sequence (such as object motion, rate of new objects entering frame or average number of frames for which an average object is visible).

\section{Experimental Setup}
\label{sec:experiments}
To illustrate the ease at which tracking by detection methods can be extended to LBT methods, we extend two popular fast multiple object trackers, SORT~\cite{bewley2016simple} and KIOU~\cite{kiout} in Section \ref{sec:setup}. These extended trackers are referred to as LBT-KIOU and LBT-SORT. Each extended tracker is evaluated on two established multiple object tracking datasets, UA-DETRAC (described in Section \ref{sec:Detrac}) and MOT20 (described in Section \ref{sec:MOT}).

\subsection{Tracker Extension}
\label{sec:setup}
For all evaluations, detections are generated using a CNN-based Retinanet object detector with ResNet50-FPN backbone trained using training set images \cite{Lin_2017_ICCV}. Likewise, localization is performed using the same CNN structure, but the localizer is trained separately on crops of size ($C \times C$). 

To use LBT, a method to select from among candidate bounding boxes for each crop is needed (Step 3a). For LBT-extended SORT, the best box is defined as the box that maximizes a score based on Euclidean distance (as used by SORT for matching) relative to the \textit{a priori} object $i$ converted to local coordinates (written as $\text{b}\widetilde{\text{o}}\text{x}_i^l:=[\tilde{x}_i^l,\tilde{y}_i^l,\tilde{w}^l_i,\tilde{h}_i^l]$). This is computed as:
\begin{equation}
    \begin{split}
    & \text{score}(\text{box}^l_{i,j},\text{b}\widetilde{\text{o}}\text{x}^l_{i}) =   \\
    & D \times\text{conf}_{i,j} - \sqrt{{(x^l_{i,j} - \tilde{x}^l_{i})^2 + (y^l_{i,j} - \tilde{y}^l_{i})^2}}, 
    \end{split}
\end{equation} 
where $D$ is a parameter that balances the confidence and Euclidean distance. %The box with the highest score is selected as the detected $\text{box}^l_{i}$ for object $i$.

For LBT-extended KIOU, intersection-over-union (IOU) similarity is used (as is used by KIOU for matching) to score candidate boxes for \textit{a priori} object $i$. The score is computed as:
\begin{equation}
    \text{score}(\text{box}^l_{i,j},\text{b}\widetilde{\text{o}}\text{x}^l_{i}) =   W \times \text{conf}_{i,j} + \Phi(\text{box}^l_{i,j},\text{b}\widetilde{\text{o}}\text{x}^l_{i}),
\end{equation}
    where $\Phi$ is a function that computes the IOU similarity between two boxes and $W$ is a scalar used to balance the two terms. The bounding box with the highest score is selected as the detected $\text{box}^l_{i}$ for object $i$. 

A number of small enhancements similar to those detailed in \cite{bewley2016simple} and \cite{bochinski2017high} are implemented. A full description of these enhancements is given in Appendix I. All parameter settings for all experiments are listed in Appendix II.

\subsection{UA-DETRAC}
\label{sec:Detrac}
The UA-DETRAC Benchmark Suite contains 10 hours of video containing traffic sequences divided into 60 training and 40 testing videos. The training and test data contain an average of 7.1 and 12.0 objects per frame, respectively \cite{wen2015ua}. LBT-extended trackers are evaluated according to the PR-metrics defined in \cite{wen2015ua} which evaluate tracking performance at varying levels of detection confidence levels. Each tracker is evaluated on the training dataset with varying numbers of frames between detection $d$. Tracking is performed at $d = 0,1,3,7,15$ and $31$ frames. Note that $d=0$ is the baseline (non-extended tracker) because detection and matching is performed every frame according to the protocol of the tracker.  We then evaluate the LBT-extended trackers  on the UA-DETRAC test data and compare to state-of-the-art algorithms, with $d$ fixed at the value which yields the best performance (in terms of PR-MOTA) on the training data.

\subsection{MOT20}
\label{sec:MOT}
The 2020 Multiple Object Tracking Benchmark Challenge (MOT20) consists of 8 sequences of pedestrian motion in crowded settings. The data is divided into 4 training and 4 testing sequences with an average of 150 and 171 objects per frame, respectively \cite{MOTChallenge20}. LBT-extended trackers are evaluated according to the CLEAR MOT metrics \cite{bernardin2008evaluating}.  Each tracker is evaluated on the training dataset with varying numbers of frames between detection $d$. Tracking is performed at $d = 0,1,3,7,15$ and $31$ frames. Again, $d=0$ is the baseline (non-extended tracker). LBT-extended trackers are then evaluated on the MOT20 evaluation server on test data and compare to state-of-the-art algorithms with $d$ fixed at the value which yields the best performance (in terms of MOTA) on the training data.

\section{Results}
\label{sec:results}
The results of LBT-extended trackers on the UA-DETRAC training dataset and the UA-DETRAC testing dataset are reported in Section \ref{sec:detrac-results}. Results on the MOT20 training dataset and the MOT20 test set are reported in Section \ref{sec:results-mot} LBT-extended trackers achieved competitive performance on all datasets, establishing a new state-of-the-art for the UA-DETRAC dataset. All tests are carried out on a single GPU.

\subsection{UA-DETRAC}
\label{sec:detrac-results}

\subsubsection{UA-DETRAC training results}
First we report the results of LBT-extended KIOU and LBT-extended SORT, evaluated on the UA-DETRAC training dataset. Tables \ref{tab:UA-train} and \ref{tab:UA-train-sort} contain abbreviated multiple object tracking results from UA-DETRAC evaluated at varying numbers of frames $d$ between detection using K-IOU and SORT, respectively, as the base tracker. $d=0$ is the baseline (detection is performed every frame). PR-metrics are calculated by performing tracking at varying detector confidence thresholds and averaging results over all confidence thresholds. Full results are reported in Appendix~IV. 

\begin{table}
\centering
\resizebox{\columnwidth}{!}{

\begin{tabular}{@{}ccccccc@{}}
\toprule
\multicolumn{1}{c}{\textbf{d}} &
\multicolumn{1}{c}{\textbf{FPS $\uparrow$}} & 
\multicolumn{1}{c}{\textbf{PR-MOTA $\uparrow$}} &
\multicolumn{1}{c}{\textbf{PR-MOTP $\uparrow$}} &
\multicolumn{1}{c}{\textbf{PR-MT $\uparrow$}} &  
\multicolumn{1}{c}{\textbf{PR-FP $\downarrow$}} &
\multicolumn{1}{c}{\textbf{PR-FN $\downarrow$}}  \\ \midrule
                                                                     
0           & 22.8          & 63.8          & 75.8          & \textbf{84.5}\% & 118486         & \textbf{81394} \\
1           & 26.5          & \textbf{67.0} & \textbf{79.1} & 69.4\%          & \textbf{55407} & 124956 \\
\textbf{3}  & 28.6          & 64.9          & 77.5          & 66.0\%          & 56665          & 134959 \\
7           & 31.4          & 62.2          & 76.3          & 59.5\%          & 58076          & 149884 \\
15          & 32.6          & 58.9          & 75.7          & 48.7\%          & 56465          & 171217 \\
31          & \textbf{34.9}          & 53.6          & 75.3          & 34.4\%          & 52338          & 209454 \\ \bottomrule
\end{tabular}
}
\caption{Abbreviated tracking metrics for UA-DETRAC training dataset with K-IOU as base tracker. Best results for each metric are shown in bold. Full results reported in Appendix IV.}
\label{tab:UA-train}
\end{table}

 As seen in Table \ref{tab:UA-train}, LBT-KIOU achieves increased accuracy (PR-MOTA) and increased frame-rate relative to the base tracker. When the number of frames between detection is small ($d=1$ and $d=3$), LBT increases the overall accuracy (PR-MOTA) of KIOU by drastically reducing the number of false positives (PR-FP). Average tracking precision (PR-MOTP) is also increased, meaning more tracklets output by the tracker correspond to ground truth objects. LBT also results in a speedup relative to the base tracker. At $d = 3$, a 25\% speedup relative to baseline is achieved in addition to an increase in accuracy. When $d=7$ a 38\% speedup is achieved for only a 1.6\% loss in PR-MOTA relative to the base tracker. Based on these results, we select $d=3$ as the best parameter setting for LBT-KIOU as it results in the largest speedup while still increasing accuracy in terms of PR-MOTA, and we use this setting for evaluation on the test data.

\begin{table}
\centering
\resizebox{\columnwidth}{!}{%

\begin{tabular}{@{}ccccccc@{}}
\toprule
\multicolumn{1}{c}{\textbf{d}} &
\multicolumn{1}{c}{\textbf{FPS $\uparrow$}} & 
\multicolumn{1}{c}{\textbf{PR-MOTA $\uparrow$}} &
\multicolumn{1}{c}{\textbf{PR-MOTP $\uparrow$}} &
\multicolumn{1}{c}{\textbf{PR-MT $\uparrow$}} &  
\multicolumn{1}{c}{\textbf{PR-FP $\downarrow$}} &
\multicolumn{1}{c}{\textbf{PR-FN $\downarrow$}}  \\ \midrule
                                                                                                                                    
0 & 20.0 & 54.9 & 72.3 & \textbf{84.0\%} & 170264 & \textbf{82785} \\ 
1 & 24.9 & \textbf{61.0} & 75.9 & 63.9\% & 73046 & 140296 \\
3 & 28.7 & 57.6 & 75.2 & 50.7\% & 64705 & 168902 \\
\textbf{7} & 32.4 & 56.0 & \textbf{76.0} & 42.3\% & 52745 & 191032 \\
15 & 33.3 & 53.3 & 75.9 & 35.1\% & 48290 & 212831 \\
31 & \textbf{34.6} & 48.4 & 75.5 & 26.8\% & \textbf{44220} & 247135 \\  \bottomrule
\end{tabular}
}
\caption{Tracking Metrics for UA-DETRAC training dataset with SORT as base tracker. Best results for each metric are shown in bold. Full results reported in Appendix IV.}
\label{tab:UA-train-sort}
\end{table}

\begin{table*}[htb]
\centering
\resizebox{\textwidth}{!}{%

\begin{tabular}{@{}lrrrrrrrr@{}} \toprule
\textbf{Tracker} &
\textbf{PR-MOTA} $\uparrow$ & 
\textbf{PR-MOTP} $\uparrow$ & 
\textbf{PR-MT}   $\uparrow$ & 
\textbf{PR-ML} $\downarrow$ & 
\textbf{PR-IDs*} $\downarrow$ & 
\textbf{PR-FM*} $\downarrow$ & 
\textbf{PR-FP*}$\downarrow$ & 
\textbf{PR-FN*}$\downarrow$ \\ \midrule
GOG               & 23.9 / 11.7                          & 47.4 / 34.4                          & 20.5\% / 10.8\%                    & 21.0\% / 21.1\%                    & 0.0158 / 0.0124                     & 0.0148 / 0.0119                    & 0.119 / 0.123                      & 0.70 / 0.70                        \\
IOUT              & 34.0 / 16.4                          & 37.8 / 26.7                          & 27.9\% / 14.8\%                    & 20.4\% / 18.2\%                    & 0.0109 / 0.0084                     & 0.0115 / 0.0089                    & \textbf{0.031 / 0.061}             & 0.64 / 0.66                        \\
JTEGCTD           & 28.4 / 14.2                          & 47.1 / 34.4                          & 23.1\% / 13.5\%                    & 18.3\% / 18.7\%                    & 0.0013 / 0.0020                     & 0.0050 / 0.0065                    & 0.096 / 0.127                      & 0.63 / 0.65                        \\
JDTIF             & - / 28.0                             & - / 41.8                             & - / 34.2\%                         & - / 20.9\%                         & - / 0.0034                          & - / 0.0166                         & - / 0.270                          & - / 0.73                           \\
MFOMOT            & 34.6 / 14.8                          & 46.6 / 35.6                          & 30.2\% / 11.9\%                    & 12.0\% / 20.8\%                    & 0.0040 / 0.0042                     & 0.0091 / 0.0098                    & 0.073 / 0.103                      & 0.52 / 0.73                        \\
KIOU              & 40.1 / 31.0                          & 49.8 / 49.9                          & 42.3\% / 37.4\%                    & \textbf{5.8\% / 10.4\%}            & 0.0021 / 0.0035                     & 0.0024 / 0.0048                    & 0.165 / 0.253                      & \textbf{0.25} / 0.46              \\
V-IOU             & 37.9 / 29.0                          & 41.7 / 35.8                          & 38.1\% / 30.1\%                    & 24.7\% / 22.2\%                    & \textbf{0.0004 / 0.0007}            & \textbf{0.0008 / 0.0012}           & 0.073 / 0.069                      & 0.66 / 0.70                        \\
DMC               & - / 14.6                             & - / 34.1                             & - / 11.6\%                         & - / 20.6\%                         & - / 0.0044                          & - / 0.0062                         & - / 0.078                          & - / 0.68                           \\
GMMA              & − / 12.3                             & − / 34.3                             & - / 10.8\%                         & - / 21.0\%                         & - / 0.0030                          & - / 0.0117                         & - / 0.124                          & - / 0.70                           \\
SCTrack-3L        & 25.9 / 12.1                          & 47.2 / 35.0                          & 15.0\% / 7.7\%                     & 20.6\% / 24.8\%                    & 0.0017 / 0.0018                     & 0.0062 / 0.0046                    & \textbf{0.047 / 0.040}             & 0.74 / 0.79                        \\ \midrule
LBT-KIOU (Ours)        & \textbf{64.5 / 46.4}                 & \textbf{79.3 / 69.5}                 & \textbf{50.1\% / 41.1\%}           & 8.2\% / 16.3\%                     & 0.0028 / 0.0051                     & 0.0091 / 0.0186                    & 0.061 / 0.113                      & 0.26 / \textbf{0.44}               \\
LBT-SORT (Ours)          & 50.7 / 35.9                          & 77.2 / 68.4                          & 31.8\% / 24.3\%                    & 17.5\% / 26.1\%                    & 0.0048 / 0.0084                     & 0.0097 / 0.0162                    & 0.057 / 0.096                      & 0.40 / 0.56                       \\ \bottomrule
\end{tabular}
}
\caption{Tracking metrics for UA-DETRAC Test Data (Beginner/Advanced) partitions. A $-$ indicates the result is not available. Results taken from \cite{lyu2018ua}. Best results for each metric are shown in bold. }
\label{tab:UA-test}
\hspace{-15.5em}\small\textsuperscript{* PR-IDs,PR-Frag,PR-FP and PR-FN are normalized by the total number of ground truth object detections across the data. See Appendix III.}
\end{table*}

As shown in Table \ref{tab:UA-train-sort}, LBT also increases the overall accuracy (PR-MOTA) of SORT, with a maximum increase of 5.1\% PR-MOTA when $d=1$. Tracking precision (PR-MOTP) is highest when $d=7$. When $d=7$, LBT-SORT is also 60\% faster than SORT and 1.1\% more accurate (PR-MOTA). Thus we select $d=7$ as the best setting for the number of frames between detection and use this setting for evaluation on the test data.

\subsubsection{UA-DETRAC testing results}
Next, we present the result of each tracker, extended with LBT, as compared to the state of the art methods for the UA-DETRAC Benchmark test dataset as reported in \cite{lyu2018ua}. Table \ref{tab:UA-test} shows that both LBT-extended SORT and LBT-extended KIOU outperform all existing methods both on overall accuracy (PR-MOTA) and tracking precision (PR-MOTP). Additionally, LBT-KIOU performs best overall in terms of mostly tracked objects (50.1\% and 41.1\% PR-MT on beginner and advanced subsets, respectively.) Notably, LBT-KIOU has the lowest rate of false negatives (PR-FN) of any tracker on the advanced test dataset partition, and the second lowest PR-FN rate on the beginner partition. This means that, despite missing some new objects as they appear, LBT-KIOU tracks known objects accurately enough to produce very few false negatives, more than making up for missed new objects.

LBT-extended KIOU establishes a new state-of-the-art for this benchmark. Further, LBT-KIOU  process at an average of 29.1 fps (realtime for this 25 fps benchmark) including the time taken to generate bounding boxes, and LBT-SORT process at an average of 28.3 fps.

\subsection{MOT20}
\label{sec:results-mot}
We now report results on MOT20. The object density and overlap in this dataset pose a large challenge for the selected base trackers KIOU and SORT because neither makes use of image information for object association. The reported results demonstrate that even in  challenging conditions for base trackers, LBT-extended trackers can still achieve competitive performance.

\subsubsection{MOT20 training results}
We report the results of LBT-extended KIOU and SORT on the MOT20 training dataset. Abbreviated results for LBT-extended KIOU and SORT on the training set of MOT20 are reported in Tables~\ref{tab:MOT-trainKIOU} and~\ref{tab:MOT-SORT}, respectively. Full results are included in Appendix V.

\begin{table}
\centering
\resizebox{\columnwidth}{!}{%
\begin{tabular}{@{}cccccc@{}}
\toprule
\multicolumn{1}{l}{\textbf{d}} & 
\multicolumn{1}{l}{\textbf{Hz} $\uparrow$} & 
\multicolumn{1}{l}{\textbf{Speedup} $\uparrow$} &
\multicolumn{1}{l}{\textbf{MOTA} $\uparrow$} & 
\multicolumn{1}{l}{\textbf{MOTP} $\uparrow$} &
\multicolumn{1}{l}{\textbf{MT} $\uparrow$}   \\ \midrule
0          & 3.7           & 0\%            & 76.1          & 74.8          & 54.8\% \\
\textbf{1} & 5.6           & 50\%           & \textbf{76.5} & 76.3          & \textbf{61.3}\% \\
3          & 7.8           & 108\%          & 74.6          & \textbf{76.6} & 57.0\% \\
7          & 10.2         & 172\%          & 70.2          & 76.5          & 47.0\% \\
15         & 12.2          & 227\%          & 65.0          & 76.3          & 35.9\% \\
31         & \textbf{12.4} & \textbf{233}\% & 58.2          & 76.2          & 24.8\% \\ \bottomrule
\end{tabular}
}
\caption{Abbreviated tracking metrics for MOT20 training data using K-IOU as base tracker. Best results for each metric are shown in bold.}
\label{tab:MOT-trainKIOU}

\end{table}

For LBT-KIOU, Table \ref{tab:MOT-trainKIOU} shows that the highest accuracy (MOTA) and largest proportion of mostly tracked objects (MT) is reported when $d=1$, and this setting is associated with a 50\% speedup in tracking relative to the base KIOU tracker ($d=0$). For a modest 1.5\% loss in MOTA ($d=3$) a 108\% speedup can be achieved. Additionally, at this setting LBT increases the tracking precision (MOTP) of the tracker by 1.8\% relative to baseline ($d = 0$). We thus select $d=1$ to be the optimal number of frames between detection.

\begin{table}
\centering
\resizebox{\columnwidth}{!}{%
\begin{tabular}{@{}cccccc@{}}
\toprule
\multicolumn{1}{l}{\textbf{d}} & 
\multicolumn{1}{l}{\textbf{Hz} $\uparrow$} & 
\multicolumn{1}{l}{\textbf{Speedup} $\uparrow$} &
\multicolumn{1}{l}{\textbf{MOTA} $\uparrow$} & 
\multicolumn{1}{l}{\textbf{MOTP} $\uparrow$} &
\multicolumn{1}{l}{\textbf{MT} $\uparrow$}   \\ \midrule
\textbf{0}  & 2.5             & 0\%            & \textbf{68.9} & \textbf{86.8} & \textbf{43.5}\% \\
1           & 4.4             & 79\%           & 64.1          & 84.8          & 33.5\%          \\
3           & 6.7             & 172\%          & 63.6          & 82.8          & 29.9\%          \\
7           & 8.7             & 254\%          & 61.0          & 81.4          & 25.9\%          \\
15          & 11.6            & 371\%          & 55.2          & 80.8          & 20.1\%          \\
31          & \textbf{14.6}   & \textbf{494}\% & 48.7          & 80.7          & 15.5\%          \\ \bottomrule 
\end{tabular}
}
\caption{Abbreviated tracking metrics for MOT20 training data using SORT as base tracker. Best results for each metric are shown in bold.}
\label{tab:MOT-SORT}
\end{table}

\begin{table*}[htb]
\centering
\resizebox{\textwidth}{!}{%
\begin{tabular}{@{}ccccccccccccccc@{}}
\toprule
\multicolumn{1}{l}{\textbf{Tracker}} & 
\multicolumn{1}{l}{\textbf{Rank}} & 
\multicolumn{1}{l}{\textbf{Hz*} $\uparrow$} &
\multicolumn{1}{l}{\textbf{MOTA} $\uparrow$} & 
\multicolumn{1}{l}{\textbf{MOTP} $\uparrow$} &
\multicolumn{1}{l}{\textbf{MT} $\uparrow$} & 
\multicolumn{1}{l}{\textbf{ML} $\downarrow$} & 
\multicolumn{1}{l}{\textbf{FP} $\downarrow$} & 
\multicolumn{1}{l}{\textbf{FN} $\downarrow$} & 
\multicolumn{1}{l}{\textbf{Recall} $\uparrow$} & 
\multicolumn{1}{l}{\textbf{Precision} $\uparrow$} & 
\multicolumn{1}{l}{\textbf{FAF} $\downarrow$} & 
\multicolumn{1}{l}{\textbf{IDs} $\downarrow$} & 
\multicolumn{1}{l}{\textbf{Frag} $\downarrow$} \\ \midrule
AGNNTrack & 1 & 16.9 & 70.8 & 79.8 & 830 (66.8) & 135 (10.9) & 47,397 & 101,754 & 80.3 & 89.8 & 10.6 & 2,106 & 2,678 \\
UniMOT & 2 & 6.0 & 70.3 & 78 & 700 (56.4) & 155 (12.5) & 25,361 & 125,631 & 75.7 & 93.9 & 5.7 & 2,581 & 6,657 \\
ort\_track & 3 & 19.2 & 69.3 & 77.6 & 765 (61.6) & 181 (14.6) & 31,790 & 124,468 & 75.9 & 92.5 & 7.1 & 2701 & 5370 \\
MOTSOT & 4 & 224.0 & 68.6 & 79.5 & 806 (64.9) & 120 (9.7) & 57,064 & 101,154 & 80.5 & 87.9 & 12.7 & 4,209 & 7,568 \\
GSDT & 5 & 0.9 & 67.1 & 79.1 & 660 (53.1) & 164 (13.2) & 31,913 & 135,409 & 73.8 & 92.3 & 7.1 & 3,131 & 9,875 \\
CSTrack & 6 & 4.5 & 66.6 & 78.8 & 626 (50.4) & 192 (15.5) & 25,404 & 144,358 & 72.1 & 93.6 & 5.7 & 3,196 & 7,632 \\
XJTU\_priv & 7 & 33.0 & 65.4 & 74.7 & 629 (50.6) & 180 (14.5) & 25,308 & 149,224 & 71.2 & 93.6 & 5.7 & 4,526 & 9,263 \\
FBMOT & 8 & 9.0 & 63.5 & 77.4 & 810 (65.2) & 122 (9.8) & 82,835 & 102,772 & 80.1 & 83.3 & 18.5 & 3,203 & 5,206 \\
Fair & 9 & 13.2 & 61.8 & 78.6 & 855 (68.8) & 94 (7.6) & 103,440 & 88,901 & 82.8 & 80.6 & 23.1 & 5,243 & 7,874 \\
LBT-KIOU (Ours) & 10 & 7.5* & 61.1 & 77.6 & 530 (42.7) & 211 (17.0) & 32,052 & 161,872 & 68.7 & 91.7 & 7.2 & 7,330 & 6,528 \\ \bottomrule
\end{tabular}
}
\caption{Tracking metrics for MOT20 Train Data. Results taken from MOT20 evaluation website \cite{MOTChallenge20}. * Hz refers to tracking time excluding detection and is not directly comparable across algorithms due to different evaluation hardware. Moreover, we report total time including end-to-end detection and tracking as we are unable to fairly separate the time of bounding box generation from other portions of our method.}
\label{tab:MOT-test}
\end{table*}

For LBT-SORT, Table \ref{tab:MOT-SORT} shows that the highest accuracy (MOTA) is reported for the baseline tracker ($d=0$) on MOT20. Even so, with a relatively modest 5.3\% loss in MOTA, a 172\% speedup can be achieved ($d=3$). Depending on use context, this may be a desirable trade-off. On this dataset, baseline SORT (68.9\% MOTA) is worse than baseline KIOU (76.1\%), demonstrating that Euclidean distance is a poor metric for matching objects in scenes with significant object overlap. By extension, Euclidean distance is also a poor metric for selecting the best bounding box from among localizer outputs. These results demonstrate that on datasets with dense objects, LBT is most useful (increases rather than decreases accuracy in addition to increasing speed) when more nuanced methods are used to select from amongst the localizer outputs.

\subsubsection{MOT20 testing results}
Lastly, we present results of LBT-extended KIOU on the MOT20 test partition, reported in Table \ref{tab:MOT-test}. Tracking outputs are evaluated on the MOT20 benchmark evaluation server.

\begin{figure}
    \centering
    \includegraphics[width = \columnwidth]{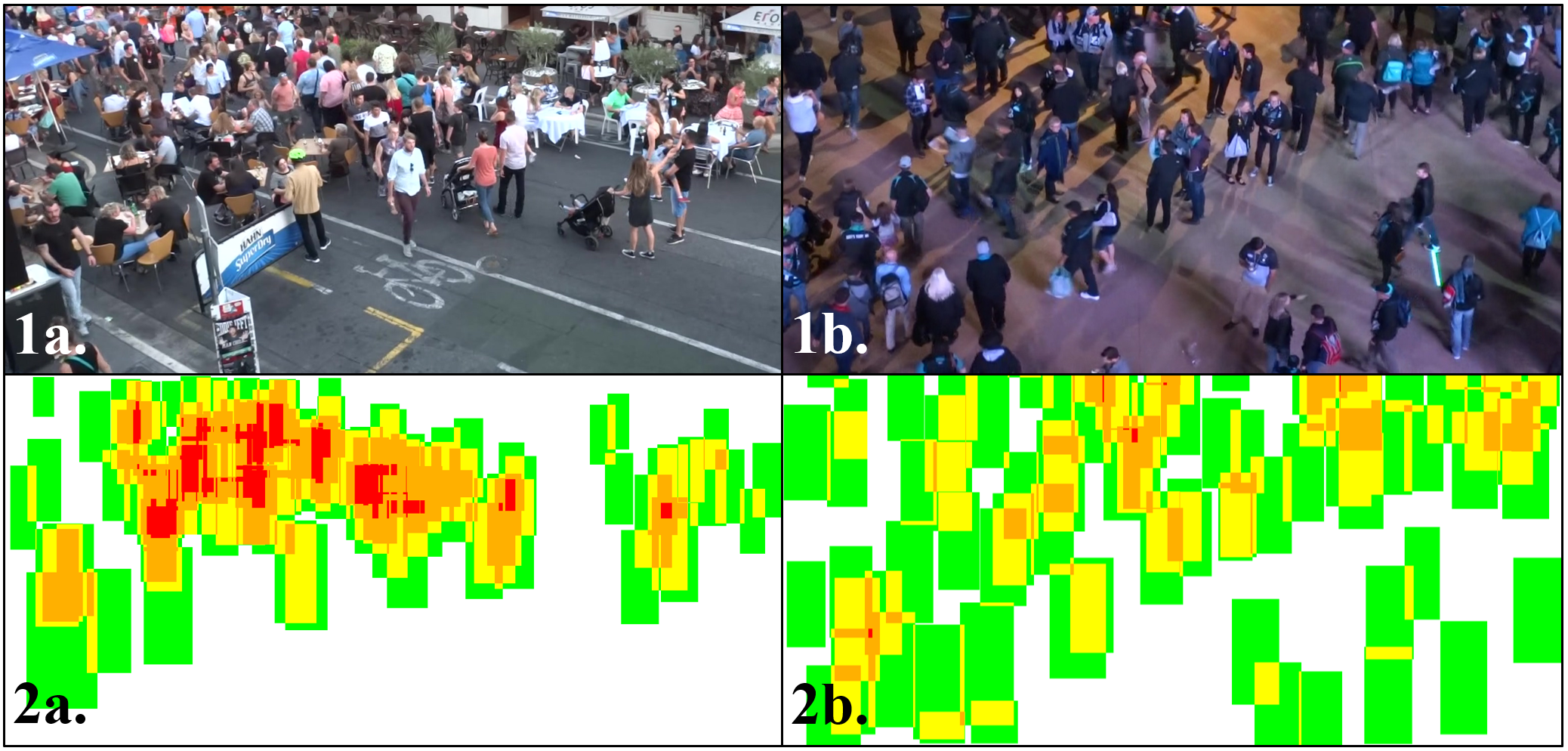}
    \caption{LBT-KIOU on test data. Example frames from  \textbf{(1a)} a sequence with high object overlap (MOTA 41.7\%) and \textbf{(1b)} a sequence with low object overlap (MOTA 77.8\%). Both frames contain approximately the same number of objects. \textbf{(2a)} and \textbf{(2b)} Object overlap in the above frames based on detected object bounding boxes. Regions of the frame containing 1 object (green), 2 overlapping objects (yellow), 3-4 overlapping objects (orange) and 5 or more overlapping objects (red) are shown.}
    \label{fig:MOT20}
\end{figure}

As of submission, LBT-KIOU places 10th overall (in terms of MOTA), even though the base tracker KIOU is not well-suited to tracking in videos with a high degree of object overlap (see Figure \ref{fig:MOT20}). The method also has high tracking precision (91.7\% MOTP), as well as a fairly low false alarm rate (FAF 7.2, 6th from among the top 10). 

Two of the four test sequences from the MOT20 test dataset contain objects that are extremely dense and overlap significantly with one another. Figure \ref{fig:MOT20} shows example frames from low and high overlap sequences. Regions of high overlap represent a challenge for KIOU (and by extension, LBT-KIOU) because KIOU relies solely on object overlap to select the most promising detection to match to a tracklet. This challenge is reflected in the high number of identity switches (IDs) of LBT-KIOU, as well as the poor MOTA on these sequences. Importantly, this is a limitation inherent to KIOU, but not LBT. A tracker that utilizes visual information in matching objects could be used as the base tracker for LBT, in which case accuracy would likely improve considerably on tracks with high overlap.

\section{Conclusion}
\label{sec:future}
This work presented \textit{Localization-based Tracking}, a powerful technique for extending existing object trackers to boost detection and tracking speed. In many cases, LBT also increases tracking accuracy considerably relative to the baseline tracker. This tracking extension was evaluated on  two popular object trackers (KIOU and SORT) and on two multiple object tracking datasets. On the UA-DETRAC dataset, Both LBT-SORT and LBT-KIOU achieved state of the art performance while producing tracklets from raw video at 28+ fps. On the MOT20 dataset, LBT increased detection and tracking speed of SORT considerably (172\% speedup with 5.3\% loss in MOTA) and increased both speed and accuracy for KIOU (50\% speedup and 0.4\% increase in MOTA). Our LBT-KIOU tracker placed 10th overall (at publication time) on the competitive MOT20 evaluation benchmark despite the inherent limitations of the base tracker KIOU on this benchmark.

This work establishes the potential of LBT to push the state of the art for object detection and tracking in realtime applications. Future work will use LBT to extend trackers that utilize visual information for object association, and will also evaluate LBT-extended trackers on benchmarks where this method excels relative to traditional tracking methods, such as scenes with low object density relative to overall frame size.

%------------------------------------------------------------------------

{\small
\bibliographystyle{ieee_fullname}
\bibliography{sources}
}

\end{document}